\documentclass{article}

\usepackage{arxiv}

\usepackage[utf8]{inputenc} 
\usepackage[T1]{fontenc}    
\usepackage{hyperref}       
\usepackage{url}            
\usepackage{booktabs}       
\usepackage{amsfonts}       
\usepackage{nicefrac}       
\usepackage{microtype}      
\usepackage{lipsum}		
\usepackage{graphicx}
\usepackage{natbib}
\usepackage{doi}
\usepackage{algorithm}
\usepackage{algorithmicx}
\usepackage{algpseudocode}
\usepackage{amsmath}
\usepackage{multirow}

\title{Clustering of Bank Customers using LSTM-based encoder-decoder and Dynamic Time Warping}

\author{
    {\includegraphics[scale=0.06]{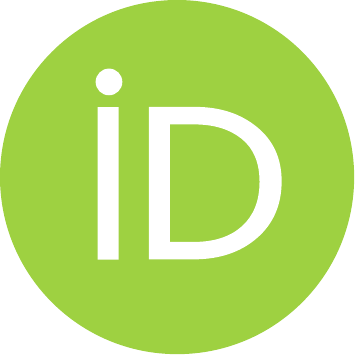}\hspace{1mm}Ehsan Barkhordar} \\
	Department of Mathematics and Computer Science\\
	Amirkabir University of Technology\\
	Tehran, Iran\\
	\texttt{bar.ehsan@aut.ac.ir} \\
	\And
	{\includegraphics[scale=0.06]{orcid.pdf}\hspace{1mm}M.~Hassan Shirali-Shahreza} \\
	Department of Mathematics and Computer Science\\
	Amirkabir University of Technology\\
	Tehran, Iran\\
	\texttt{hshirali@aut.ac.ir} \\
	\And
	{\includegraphics[scale=0.06]{orcid.pdf}\hspace{1mm}H.~Reza Sadeghi} \\
	Department of Mathematics and Computer Science\\
	Amirkabir University of Technology\\
	Tehran, Iran\\
	\texttt{sadeghihamid@aut.ac.ir} \\
}

\renewcommand{\headeright}{}
\renewcommand{\shorttitle}{Clustering of Bank Customers using LSTM-based encoder-decoder and Dynamic Time Warping}

\hypersetup{
pdftitle={Clustering of Bank Customers using LSTM-based encoder-decoder and Dynamic Time Warping},
pdfsubject={Clustering, Deep learning, Banking},
pdfauthor={Ehsan Barkhordar, M.~Hassan Shirali-Shahreza, H.~Reza Sadeghi},
pdfkeywords={Customer clustering, Time-series clustering, LSTM encoder-decoder, Dynamic time warping},
}

\begin{document}
\maketitle

\begin{abstract}
	Clustering is an unsupervised data mining technique that can be employed to segment customers. The efficient clustering of customers enables banks to design and make offers based on the features of the target customers. The present study uses a real-world financial dataset (Berka, 2000) to cluster bank customers by an encoder-decoder network and the dynamic time warping (DTW) method. The customer features required for clustering are obtained in four ways: Dynamic Time Warping (DTW), Recency Frequency and Monetary (RFM), LSTM encoder-decoder network, and our proposed hybrid method. Once the LSTM model was trained by customer transaction data, a feature vector of each customer was automatically extracted by the encoder.
    Moreover, the distance between pairs of sequences of transaction amounts was obtained using DTW. Another vector feature was calculated for customers by RFM scoring. In the hybrid method, the feature vectors are combined from the encoder-decoder output, the DTW distance, and the demographic data (e.g., age and gender). Finally, feature vectors were introduced as input to the k-means clustering algorithm, and we compared clustering results with Silhouette and Davies–Bouldin index. As a result, the clusters obtained from the hybrid approach are more accurate and meaningful than those derived from individual clustering techniques. In addition, the type of neural network layers had a substantial effect on the clusters, and high network error does not necessarily worsen clustering performance.
\end{abstract}


\section{Introduction}
Banks seek to obtain competitive advantages in today’s devastating competition and globalization \citep{moin2012use}. It is essential for the banking sector to identify advanced big data analysis methods, e.g., data mining techniques, in order to extract valuable information from a massive amount of data and improve strategic management and customer satisfaction \citep{hassani2018digitalisation}.

Service marketing research has shown that companies should not offer the same services for all customers in most cases. Therefore, customer clustering and customer relationship management are determinants of business survival \citep{ansari2016customer}. Efficient customer clustering enables the effective segmentation of customers. Clustering classifies customers with similar features and demands into the same group. Through customer clustering, banks can better identify customer behavior and develop more effective marketing strategies. Also, banks take a step toward data-driven decision-making by customer clustering, enhancing their knowledge of customer behavior. Clustering is typically the initial step of customer segmentation. Thus, the present work seeks to extract efficient features to cluster bank customers based on their transactions.

\section{Literature review}
Previous studies clustered customers based on customer equity through the k-means and k-medoids techniques, comparing the performances of the two approaches. They found that k-means clustering outperformed k-medoids clustering based on both the average within-cluster (AWC) distance and the Davies-Bouldin index \citep{aryuni2018customer}. A relatively recent work employed self-organizing maps and k-means to cluster customers. The variables used are grouped into three as demographical variables, categorical consumption variables, and summary consumption variables \citep{yanik2019som}. Customers were clustered based on their three-month consumption and demographic data.

Although earlier works exploited either customer transaction data or demographic data, Davood et al. \citep{dawood2019improve} utilized a combination of transaction and demographic data to obtain more accurate results. Therefore, banks can achieve their business objectives by finding different groups of customers with similar financial behavior. The present study proposes an intelligent model of bank customer clustering based on customer transactions. The proposed model converts customer transactions and demographic data (e.g., gender and age) into a vector in a latent space. This vector representation of customer data helps cluster customers with similar transaction behavior in the same group. Clusters of higher accuracy can be obtained by using vector representation and customer features of higher optimality.

\section{Theoretical background}
Transaction data refer to the dataset of an event such as a financial transaction or online payment. Each transaction involves at least a time dimension and the transition amount. Here, transactions refer to bank transactions, such as payments or money transfers.

\begin{figure} 
    \centering
    \includegraphics[scale=0.3]{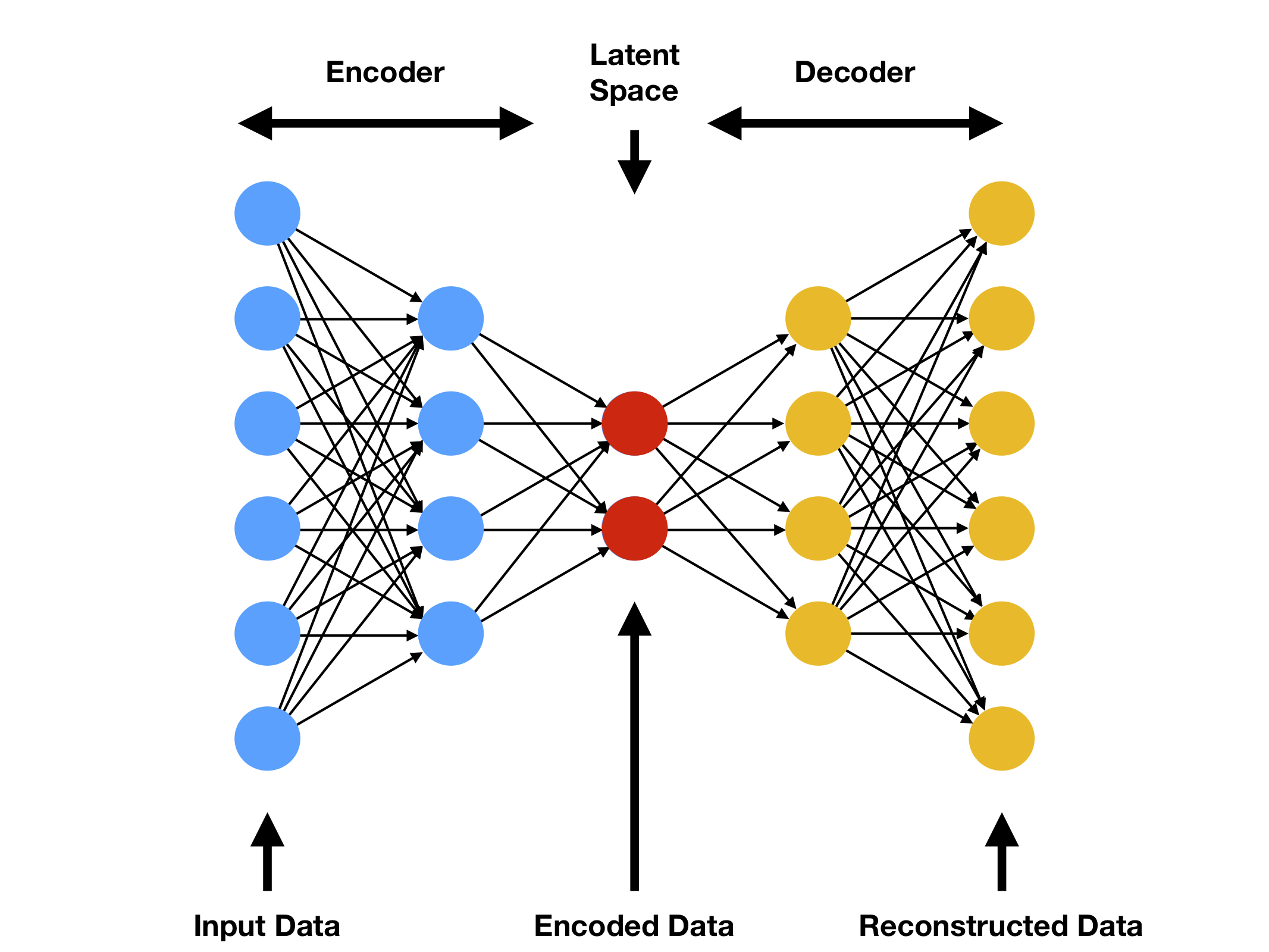}
    \caption{Architecture of encoder-decoder \citep{Steven2019}.}
    \label{fig:fig1}
\end{figure}

An artificial neural network (ANN) is a set of algorithms that attempt to detect fundamental relationships in a data set through a human brain-inspired process. An autoencoder is an unsupervised learning method in which ANNs are employed for representation learning. In particular, a neural network architecture is designed to impose a bottleneck to force a compressed representation of the primary input. Compression and reconstruction are complicated. However, a data structure (i.e., a correlation between input features) can be trained and used when forcing through the bottleneck. This group of ANNs is employed to reduce dimensionality and diminish processing time and memory costs. These concepts were introduced by Hinton (1980) and the PDP Research Group. Autoencoders were considered restricted Boltzmann machines (RBMs) for deep architecture in the 2000s.

Auto-Encoder is a neural network that attempts to reconstruct its input at its output \citep{shi2018auto}. An autoencoder consists of two parts: an encoder and a decoder (See Figure \ref{fig:fig1}), which generally are implemented by neural networks. The encoder and decoder can be viewed as two functions \(z = f(x) \) and \(r = g(z) \), the \(f(x)\) maps data point \(x\) from data space to feature space, while \(g(z)\) produces a reconstruction of data point \(x\) by mapping \(z\) from feature space to data space. In modern autoencoders, the two functions \(z = f(x) \) and \(r = g(z) \) usually are stochastic functions  \(P_{encoder}=(z|x)\) and  \(P_{decoder}=(r|z)\) , where \(r\) is the reconstruction of \(x\). From the view of applications, it is important to note that one does not wish autoencoders to simply learn to copy of the input \(x\). In other words, autoencoders are usually restricted in some ways that allow them to approximately learn the copy of the inputs \citep{zhai2018autoencoder}.

The present study also evaluated dynamic time warping (DTW). DTW measures the dependency or similarity of two time-series that may differ in time for time series. For example, DTW can find the similarity of two walking patterns, even when the waking speeds or accelerations are not the same in time intervals. DTW has analyzed time series of audio, video, and image data.

\begin{figure} 
    \centering
    \includegraphics[scale=0.5]{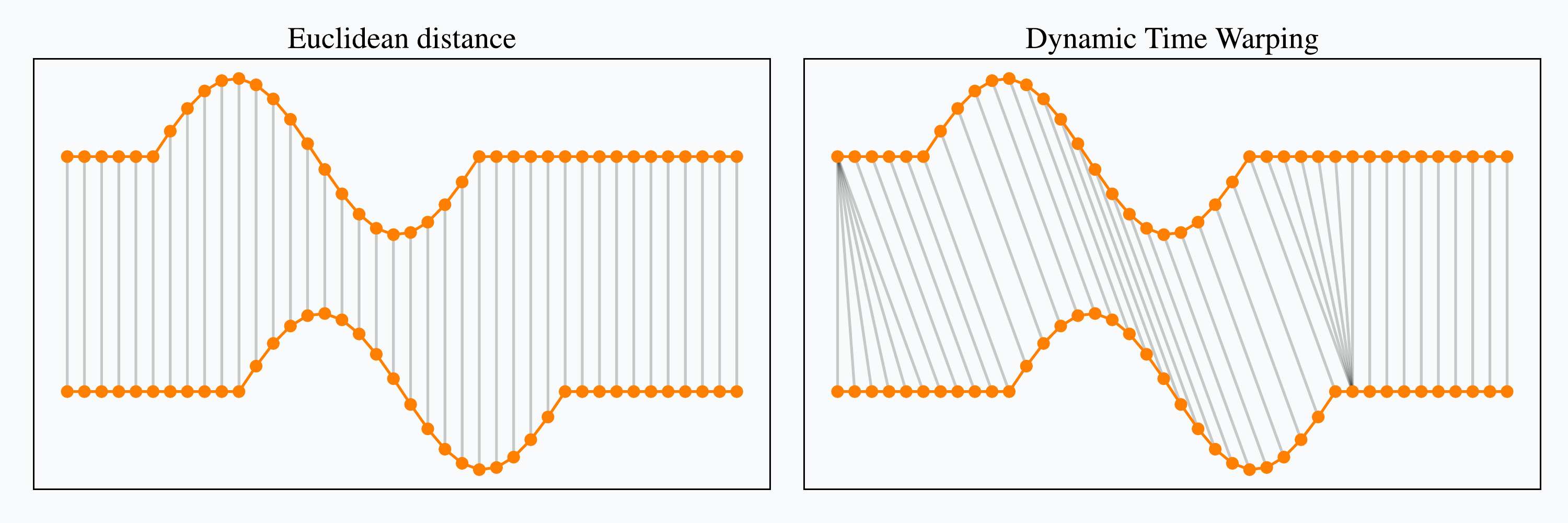}
    \caption{Comparison of Euclidean and DTW distances \citep{Romain2021}.}
    \label{fig:fig2}
\end{figure}

\algnewcommand\algorithmicinput{\textbf{Let}}
\algnewcommand\Let{\item[\algorithmicinput]}

\begin{algorithm}
\caption{Dynamic Time Warping \citep{petitjean2011global}}\label{alg:cap}
\begin{algorithmic}[1]
    \Require $A = <a_{1},\cdots,a_{S}>$
    \Require $B = <b_{1},\cdots,b_{T}>$
    \Let $\delta$ be a distance between coordinates of sequences
    \Let $m[S, T]$ be the matrix of couples (cost,path)
        \State $m[1,1] \gets\left(\delta\left(a_{1}, b_{1}\right),(0,0)\right)$
        \For{$i \gets 2$ to $S$}
            \State $m[i, 1] \leftarrow\left(m[i-1,1,1]+\delta\left(a_{i}, b_{1}\right),(i-1,1)\right)$
        \EndFor
        \For{$j \gets 2$ to $T$}
            \State $m[1, j] \leftarrow\left(m[1, j-1,1]+\delta\left(a_{1}, b_{j}\right),(1, j-1)\right)$
        \EndFor
        
        \For{$i \gets 2$ to $S$}
            \For{$j \gets 2$ to $T$}
                \State ${ minimum } \leftarrow minVal(m[i-1, j], m[i, j-1], m[i-1, j-1])$
                \State $m[i, j] \leftarrow\left(first (minimum)+\delta\left(a_{i}, b_{j}\right), second (minimum)\right)$
            \EndFor
        \EndFor
        \State \Return m[S, T]
\end{algorithmic}
\end{algorithm}

\section{Proposed customer clustering framework}
Customer data is typically stored in a raw form in the databases of banks, without labels of valuable or uncreditworthy customers. Thus, unsupervised approaches are more efficient in the extraction of customer features through transaction data.

The present study adopted a multilayer LSTM-based encoder-decoder to extract customer features from customer transactions. The input and output are two lists of the transaction sequences of customers. The output of the encoder is known as the latent space. Thus, each input transaction has a vector representation in the latent space that contains most characteristics of the transaction sequence. The LSTM layer was employed to help the network better learn the transaction time-series \citep{yu2019review}.

The performance of this encoder-decoder model was improved by incorporating the attention mechanism \citep{haghani2018audio}. This mechanism is used to tackle initial input sequence element information forgotten in the coded vector when the input sequence is long.  In each output step, the last decoder latent state is utilized to generate an attention vector in the encoder for downsizing and disseminating information from the encoder to the decoder.

Customer transactions are introduced as inputs in chronological order to the proposed neural network. To this end, transactions are classified based on the account numbers of customers. Each transaction involves four aspects: (1) type, (2) timestamp, (3) amount, and (4) account balance. In order to equalize the dimensions of transactions, two-dimensional zero arrays are added to the list of transactions with a length less than the maximum length of transactions. A matrix of all customer transactions was obtained. The following equations represent the transaction feature vector, the feature matrix of a customer, and the feature vector of all customers:

\begin{equation} \label{GrindEQ__1_} 
\left[ \begin{array}{cccc}
type(p_0) & amount(m_0) & balance(b_0) & timestamp(t_0) \end{array}
\right] 
\end{equation} 
\begin{equation} \label{GrindEQ__2_} 
\left[ \begin{array}{cccc}
p_0 & m_0 & b_0 & t_0 \\ 
p_1 & m_1 & b_1 & t_1 \\ 
\vdots  & \vdots  & \vdots  & \vdots  \\ 
{\ \ p}_{i-2} & {\ \ m}_{i-2} & {\ \ b}_{i-2} & t_{i-2}\ \  \\ 
\ \ p_{i-1} & m_{i-1} & b_{i-1} & t_{i-1\ \ } \\ 
0 & 0 & 0 & 0 \\ 
\vdots  & \vdots  & \vdots  & \vdots  \\ 
0 & 0 & 0 & 0 \end{array}
\right] 
\end{equation} 
\begin{equation} \label{GrindEQ__3_} 
\left[ \begin{array}{ccc}
\left[ \begin{array}{cccc}
p_{0,0} & m_{0,0} & b_{0,0} & t_{0,0} \\ 
p_{1,0} & m_{1,0} & b_{1,0} & t_{1,0} \\ 
\vdots  & \vdots  & \vdots  & \vdots  \\ 
p_{i-1,0} & m_{i-1,0} & b_{i-1,0} & t_{i-1,0} \\ 
0 & 0 & 0 & 0 \\ 
\vdots  & \vdots  & \vdots  & \vdots  \\ 
0 & 0 & 0 & 0 \end{array}
\right] & \dots  & \left[ \begin{array}{cccc}
p_{0,j} & m_{0,j} & b_{0,j} & t_{0,j} \\ 
p_{1,j} & m_{1,j} & b_{1,j} & t_{1,j} \\ 
\vdots  & \vdots  & \vdots  & \vdots  \\ 
p_{k-1,j} & m_{k-1,j} & b_{k-1,j} & t_{k-1,j} \\ 
0 & 0 & 0 & 0 \\ 
\vdots  & \vdots  & \vdots  & \vdots  \\ 
0 & 0 & 0 & 0 \end{array}
\right] \end{array}
\right] 
\end{equation}

The data should be normalized before training. Normalization was carried out using the z-score as:

\[Z=\frac{x-\mu }{\sigma }\]

Where Z is the final value, \textit{x} is the initial value, $\mu$ is the mean, and $\sigma$ is the standard deviation of the data. The average of the z-scores is zero. A Start-of-Sequence (SOS) label is applied to indicate the start of the transaction sequence so that the teacher forcing model is used. Teacher forcing is a strategy of training recurrent neural networks in which the output of a time step is used as the input of the next time step. The input of the decoder has only a start, and the output of the decoder has only an end. Thus, the input is shifted by a time step. Thus, it is required to use $\left[ \begin{array}{cccc}
-1 & -1 & -1 & -1 \end{array}
\right]$ for the start of the sequence. Also, $\left[ \begin{array}{cccc}
\mathrm{-2} & -2 & -2 & \mathrm{-2} \end{array}
\right]$ is used for the end of the sequence. For example, 
\[Input=\left[ \begin{array}{cccc}
-1 & -1 & -1 & -1 \\ 
p_0 & m_0 & b_0 & t_0 \\ 
p_1 & m_1 & b_1 & t_1 \\ 
\vdots  & \vdots  & \vdots  & \vdots  \\ 
\ \ p_{i-1} & m_{i-1} & b_{i-1} & t_{i-1\ \ } \\ 
0 & 0 & 0 & 0 \\ 
\vdots  & \vdots  & \vdots  & \vdots  \\ 
0 & 0 & 0 & 0 \end{array}
\right],\ \ \ Output=\left[ \begin{array}{cccc}
p_0 & m_0 & b_0 & t_0 \\ 
p_1 & m_1 & b_1 & t_1 \\ 
\vdots  & \vdots  & \vdots  & \vdots  \\ 
\ \ p_{i-1} & m_{i-1} & b_{i-1} & t_{i-1\ \ } \\ 
-2 & -2 & -2 & -2 \\ 
0 & 0 & 0 & 0 \\ 
\vdots  & \vdots  & \vdots  & \vdots  \\ 
0 & 0 & 0 & 0 \end{array}
\right]\]

Once training is completed, it is required to define the decoder as a distinct model to receive customer transaction inputs and assign a feature vector to each customer to represent helpful information on the customer's transactions. The dimensionality of the latent space is recommended to be lower than the maximum number of transactions. The output of the decoder for each customer is a feature vector as follows:
\[Custmer\ Feature\ Vector=\left[ \begin{array}{c}
f_0 \\ 
f_1 \\ 
f_2 \\ 
\vdots  \\ 
\ \ f_l \end{array}
\right]\] 
Where $f_i$ denotes feature \textit{i} in the latent representation and dimension \textit{l} is the latent layer. The final decoder output, which contains the features of all the customers, is obtained as a matrix:
\[Decoder\ Output:\left[ \begin{array}{ccc}
\left[ \begin{array}{c}
f_{0,j} \\ 
f_{1,j} \\ 
f_{2,j} \\ 
\vdots  \\ 
\ \ f_{l,j} \end{array}
\right] & \dots  & \left[ \begin{array}{c}
f_{0,j} \\ 
f_{1,j} \\ 
f_{2,j} \\ 
\vdots  \\ 
\ \ f_{l,j} \end{array}
\right] \end{array}
\right]\] 
Additionally, the DTW distance was employed to extract customer features to measure the similarity of pairs of customers. DTW obtains the distance between the transaction amounts of the two customers. Each customer's transaction amount and time are extracted and converted into a two-dimensional sequence in chronological order. Then, DTW was applied to measure the distance between the two transaction sequences. It is the minimum difference between the two transaction sequences under certain conditions. The transaction sequences of the customers are stored in a matrix whose rows and columns are bank transactions, and each entry denotes the DTW distance of the two transactions.

\begin{figure} 
    \centering
    \includegraphics[scale=0.8]{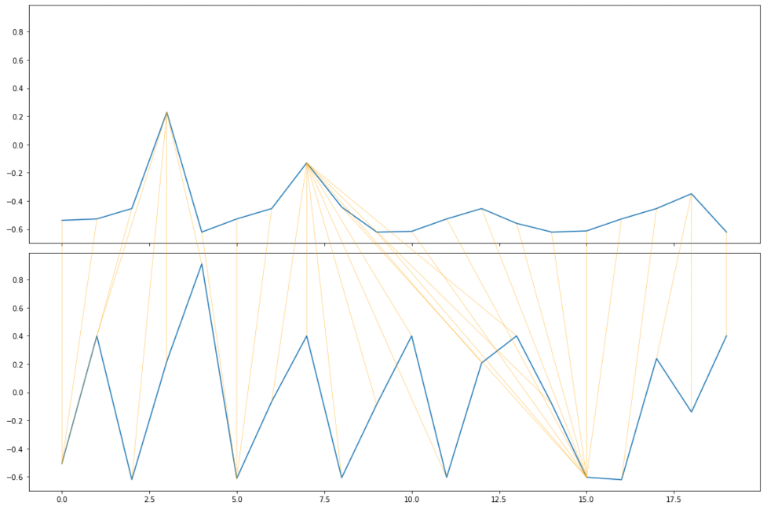}
    \caption{Optimal DTW path to calculate the distance between two customer transaction amount series.}
    \label{fig:fig3}
\end{figure}

There are as many distances as customers. These distances can be considered as a new feature vector. Then, the DWT features and LSTM features are concatenated. Moreover, the demographic features of the customers (e.g., age, gender, longitude, and latitude) are included. Thus, three feature matrices are concatenated to construct a new matrix. The concatenated matrix has a feature with a size of m+n+d for each customer, in which \textit{m} denotes the LSTM feature-length, \textit{n} denotes the DTW distance, and \textit{d} stands for the demographic data.

Indeed, it is required to reduce the dimensionality of the feature vector to improve clustering performance and visualize the results. In order to reduce feature vector dimensionality, the present study employed principal component analysis (PCA). 

Finally, the dimension-reduced feature matrix was clustered using the k-means algorithm in light of its satisfactory performance for a large amount of data. Although different augmentations of K-means clustering have been introduced, the present study adopted the elbow method to find the efficient number of customer clusters. 

\section{Results and discussion}

A shortage of publically available data due to customer privacy protection reasons was a significant challenge. The present study employed an enhanced variant of the Beka Dataset - the original database was published by Berka (2000). The Beka Database is the financial dataset of a bank in the Czech Republic. It contains data of over 5300 customers with nearly 1,000,000 transactions. Also, the bank granted 700 loans and issued approximately 900 credit cards (provided in the dataset) \citep{berka2000guide}. This study focuses on transactions and customer tables, as shown in Table.\ref{tab:table1}.

\begin{table}
	\caption{Sample customer transactions}
	\centering
	\begin{tabular}{cccccc}
		\toprule
		Trans\_ID  & Account\_ID   & Type & Amount & Balance  & Timestamp\\
		\midrule
		T00695247 & A00002378 & Credit & 700.0 & 700.0 & 1356998400     \\
		T00171812 & A00000576 & Credit & 900.0 & 900.0 & 1356998400     \\
		T01117247 & A00003818 & Credit & 600.0 & 600.0 & 1356998400     \\
		T00579373 & A00001972 & Credit & 400.0 & 400.0 & 1357084800     \\
		\bottomrule
	\end{tabular}
	\label{tab:table1}
\end{table}

\begin{table}
	\caption{Sample customer features}
	\centering
	\begin{tabular}{cccccc}
		\toprule
		Customer\_ID  & Account\_ID   & Gender & Age & Latitude  & Longitude\\
		\midrule
		C00000001 & A00000001 & 0.0 & 29 & 35.08449 & -106.65114    \\
		C00000002 & A00000002 & 1.0 & 54 & 40.71427 & -74.00597     \\
		C00000004 & A00000003 & 1.0 & 43 & 39.76838 & -86.15804     \\
		\bottomrule
	\end{tabular}
	\label{tab:table2}
\end{table}

The present study has utilized 70\% of the data as the training dataset, 20\% as the test dataset, and the remaining 10\% as the validation dataset for the encoder-decoder neural network. ReLU and sigmoid activation functions were utilized. The average loss was calculated to be 0.3509.

\begin{figure} 
    \centering
    \includegraphics*[scale=1]{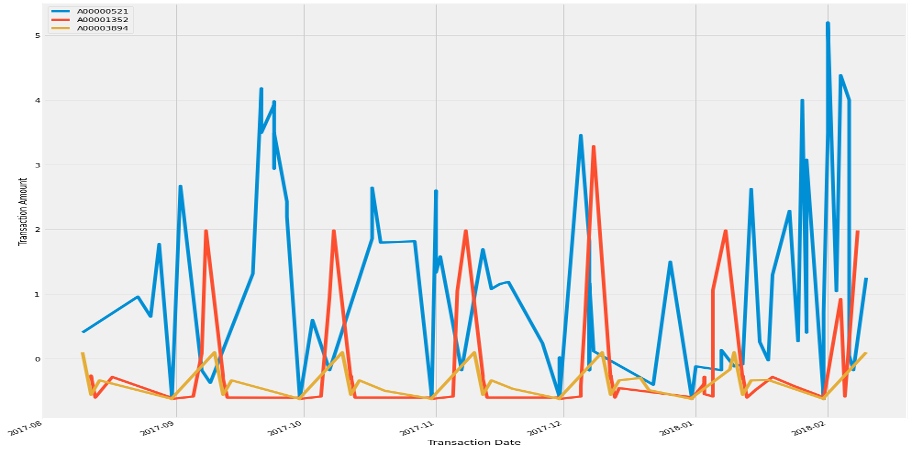}
    \caption{Comparison of the sequence of transaction amounts of three customers in three different clusters by DTW method.}
    \label{fig:fig4}
\end{figure}

Then, customer clustering was implemented based on the customer distance matrix through DTW. The customers were divided into three clusters using the k-means method. A customer was randomly selected from each cluster by searching the dataset, plotting their transaction sequences. As you see in Figure.\ref{fig:fig4} Customer Red had lower transaction amounts than Customer Blue. Moreover, Customer Yellow had a larger distance in their transactions. Therefore, it can be said that these clusters were significantly distinct.

\begin{figure} 
    \centering
    \includegraphics*[scale=0.6]{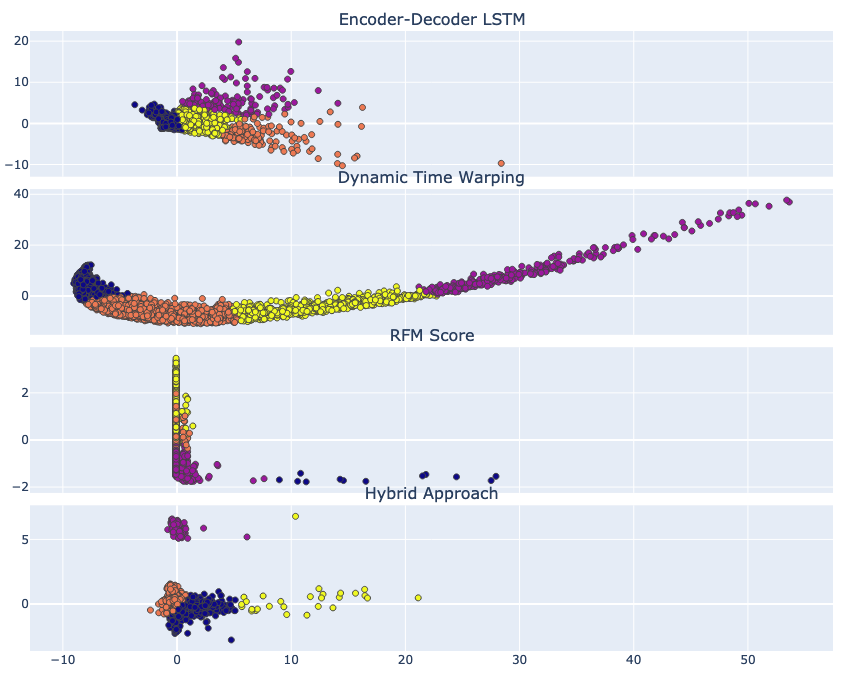}
    \caption{K-means clustering in different methods.}
    \label{fig:fig5}
\end{figure}

Once neural network learning and DTW customer distance calculations are completed, we used another traditional clustering method to compare the results better. This method, called RFM, scores users based on three values: recency, frequency, and monetary.
the customers were clustered in four scenarios:
\begin{itemize}
    \item LSTM features
    \item DTW distances
    \item RFM score
    \item Hybrid
\end{itemize}

Table.\ref{tab:table3} compares the clustering approaches. The Silhouette Coefficient (SC) and Davies-Bouldin Index (DBI) were utilized to evaluate the clusters.

\begin{table}[h]
	\caption{Comparison of the K-means clustering results}
	\centering
	\begin{tabular}{ccccccccc}
		\toprule
        \multirow{2}[3]{*}{No. of Clusters} & \multicolumn{2}{c}{Encoder-Decoder LSTM} & \multicolumn{2}{c}{Dynamic Time Warping} & \multicolumn{2}{c}{RFM Score} & \multicolumn{2}{c}{Hybrid Approach} \\
		\cmidrule(lr){2-3}\cmidrule(lr){4-5} \cmidrule(lr){6-7}  \cmidrule(lr){8-9}
		& SC  & DBI  & SC  & DBI & SC  & DBI & SC  & DBI\\
    
        \midrule
        2 & 0.691 & 0.928 &     0.586 & 0.706 &     0.495 & 0.828 &     0.870 & 0.826 \\
        3 & 0.695 & 0.796 &     0.559 & 0.602 &     0.418 & 0.904 &     0.862 & 0.381 \\
        4 & 0.508 & 0.862 &     0.572 & 0.560 &     0.437 & 0.745 &     0.761 & 0.439 \\
        5 & 0.491 & 0.787 &     0.541 & 0.605 &     0.438 & 0.745 &     0.409 & 0.608 \\
        6 & 0.498 & 0.793 &     0.533 & 0.614 &     0.430 & 0.741 &     0.421 & 0.637 \\
        \bottomrule
	\end{tabular}
	\label{tab:table3}
\end{table}

\section{Conclusion}

After analyzing the results, some points can be summarized as follows:

\begin{enumerate}
    
    \item  The attention mechanism enhanced the accuracy of the proposed neural network and this clustering performance.
    
    \item  A low error does not necessarily lead to high clustering performance -- the opposite was the case with most cases. 
    
    \item  The training and testing of the model showed that the final clusters would be more unrealistic when the dimensionality of the latent space was larger than the maximum number of transactions of a customer.
    
    \item  The DTW-extracted features had high continuity. Therefore, the individual DTW approach did not yield significant clustering performance. 
    
    \item  The proposed hybrid model was found to have higher performance evaluation indices than the two individual approaches in most cases.
    
    \item  Pre-concatenation dimensionality reduction led to higher clustering performance than post-concatenation dimensionality reduction. 
\end{enumerate}

\bibliographystyle{unsrtnat}
\bibliography{references}  






\end{document}